\documentclass{article}
\usepackage{spconf,amsmath,graphicx,color}
\usepackage{amssymb}
\usepackage{multirow,cite}
\usepackage{threeparttable}
\usepackage{epstopdf}
\graphicspath{{figs/}}

\title{Partial AUC optimization based deep speaker embeddings with class-center learning for text-independent speaker verification}
\name{Zhongxin Bai, Xiao-Lei Zhang, and Jingdong Chen
\thanks{This work was supported in part by the Key Program of National Science of Foundation of China (NSFC) under Grant No. 61831019 and
the NSFC and Israel Science Foundation (ISF) joint research program under Grant No. 61761146001.}
      }
\address{Center of Intelligent Acoustics and Immersive Communications and \\
  School of Marine Science and Technology, Northwestern Polytechnical University\\
  zxbai@mail.nwpu.edu.cn, xiaolei.zhang@nwpu.edu.cn, jingdongchen@ieee.org}
\begin{document}
\ninept
\maketitle
\begin{abstract}
Deep embedding based text-independent speaker verification has demonstrated superior performance to traditional methods in many challenging scenarios. Its loss functions can be generally categorized  into two classes, i.e., verification and identification. The verification loss functions match the pipeline of speaker verification, but their implementations are difficult. Thus, most state-of-the-art deep embedding methods use the identification loss functions with softmax output units or their variants. In this paper, we propose a verification loss function, named the maximization of partial area under the Receiver-operating-characteristic (ROC) curve (pAUC), for deep embedding based text-independent speaker verification. We also propose a class-center based training trial construction method to improve the training efficiency, which is critical for the proposed loss function to be comparable to the identification loss in performance. Experiments on the Speaker in the Wild (SITW) and NIST SRE 2016 datasets show that the proposed pAUC loss function is highly competitive with the state-of-the-art identification loss functions.
\end{abstract}


\section{Introduction}
\label{sec:intro}

Text independent speaker verification aims to verify whether an utterance is pronounced by a hypothesized speaker according to his/her pre-recorded utterances without limiting the speech contents. The state-of-the-art text-independent speaker verification systems \cite{snyder2018xvector,snyder2019speaker,xie2019utterance,villalba2019state} use deep neural networks (DNNs) to project speech recordings with different lengths into a common low dimensional embedding space where the speakers' identities are represented. Such a method is called \textit{deep embedding}, where the embedding networks have three key components---network structure \cite{snyder2018xvector,xie2019utterance,heigold2016end,jung2018complete,li2017deep}, pooling layer \cite{snyder2018xvector,zhu2018self,tang2019deep,cai2018exploring,bhattacharya2017deep,gao2019improving}, and loss function \cite{wang2019discriminative,li2019boundary,zhang2018text,wan2018generalized,mingote2019optimization}. This paper focuses on the last part, i.e., the loss functions.

Generally, there are two types of loss functions, i.e., identification and verification loss functions. The former is mainly cross-entropy loss with softmax output units or softmax variants \cite{xie2019utterance,liu2019speaker} such as ASoftmax, AMSoftmax, and ArcSoftmax whose role is to change the linear transform of the softmax function to a cosine function with a margin controlling the distance between speakers's spaces. Different from \cite{xie2019utterance,liu2019speaker}, which conducts multi-class classification from a single classifier, \cite{mingote2019optimization11} conducts one-versus-all classification with multiple binary classifiers.
In comparison, the verification loss functions mainly consist of pairwise- or triplet-based loss functions \cite{heigold2016end,zhang2018text,wan2018generalized,mingote2019optimization,novoselov2018triplet,li2017deep1}. The fundamental difference between the verification and identification loss functions is that the former needs to construct pairwise or triplet training trials, which imitates the enrollment and test stages of speaker verification. Although this imitation matches the pipeline of speaker verification ideally, its implementation faces many difficulties in practice. One of those is  that the number of all possible training trials increase cubically or quadratically with the number of training utterances. One way to circumvent this issue is through selecting part of the informative training trials that are difficult to be discriminated. 
But finding a good selection method is a challenging task. Moreover, the optimization process with the verification function is not so stable in comparison with that with the identification loss. As a result, the state-of-the-art deep embedding methods optimize the identification loss directly, or fine-tune an identification loss based DNN with the verification loss \cite{li2017deep1,Chung2018}.

Despite of some disadvantages, a great property of the verification loss functions is that the training process is consistent with the evaluation procedure, which make it more proper for speaker verification in comparison with the identification loss. In \cite{bai2019partial}, we proposed a new verification loss, named \textit{maximizing partial area under the ROC curve} (pAUC), for training back-ends. The motivation and advantages of the pAUC maximization over other verification loss functions were shown in \cite{bai2019partial}.

In this paper, we extend the work in \cite{bai2019partial}. Motivated from the advantages of the identification loss, we improve the procedure of its training trial construction in \cite{bai2019partial} by a \textit{class-center learning} method. This approach first learns the centers of classes of the training speakers, and then uses the class-centers as enrollments to construct training trials at each optimization epoch of the pAUC deep embedding. Experiments are conducted on the Speaker in the Wild (SITW) and NIST SRE 2016 datasets. Results demonstrated that the proposed pAUC deep embedding is highly competitive in performance with the state-of-the-art identification loss based deep embedding methods with the Softmax and ArcSoftmax output units. Note that a very recent work proposed at the same time as our work in \cite{bai2019partial} maximizes the area under the ROC curve (AUC) for text-dependent speaker verification \cite{novoselov2018triplet}.  It can be shown that AUC is a particular case of pAUC and experimental results show the pAUC deep embedding outperforms the AUC deep embedding significantly.

\section{Deep embedding via pAUC optimization} \label{sec:pAUC_optimization}
\subsection{Objective function}
\label{sec:obj}
\begin{figure}[t!]
  \centering
  \includegraphics[width=2.0 in]{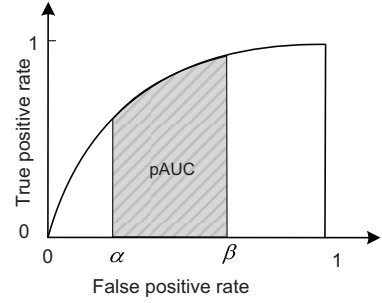}
  \caption{Illustrations of the ROC curve, AUC, and pAUC. }
  \label{fig:pAUC_schematic}
\end{figure}

In the training stage, the DNN model of the pAUC deep embedding system outputs a similarity score $s$ for a pair of input utterances. In the test stage, it outputs an embedding vector $\mathbf{h}$ from the top hidden layer for each input utterance, and uses $\mathbf{h}$ for verification.
 Because the gradient at the output layer can be transferred to the hidden layers by backpropagation, we focus on presenting the pAUC optimization at the output layer as follows:

 In the training stage, we construct a pairwise training set $\mathcal{T}=\{(\mathbf{x}_n,\mathbf{y}_n;l_n)|n=1,2,\cdots,N\}$
where $\mathbf{x}_n$ and $\mathbf{y}_n$ are the representations of two utterances at the output layer of the DNN model, and
$l_n$ is the ground-truth label indicating the similarity of $\mathbf{x}_n$ and $\mathbf{y}_n$ (if $\mathbf{x}_n$ and $\mathbf{y}_n$ come from the same speaker, $l_n=1$; otherwise, $l_n=0$). Given a soft similarity function $ f(\cdot)$, we obtain a similarity score of $\mathbf{x}_n$ and $\mathbf{y}_n$, denoted as $s_n = f(\mathbf{x}_n,\mathbf{y}_n)$ where $s_n\in \mathbb{R}$. The hard decision of the similarity of $\mathbf{x}_n$ and $\mathbf{y}_n$ is:
\begin{equation}\label{eq:similarity}
  \hat{l}_n = \left\{\begin{array}{l}
  1,\quad\mbox{if } s_n \geq \theta\\
  0,\quad \mbox{otherwise}
  \end{array}
   \right.,\forall n = 1,\ldots,N,
\end{equation}
where $\theta$ is a decision threshold. Given a fixed value of $\theta$, we are able to compute a true positive rate (TPR) and a false positive rate (FPR) from $\hat{l}_n,\forall n = 1,\ldots, N$. TPR is defined as the ratio of the positive trials (i.e. $l_n=1$) that are correctly predicted (i.e. $\hat{l}_n = 1$) over all positive trials. FPR is the ratio of the negative trials (i.e. $l_n=0$) that are wrongly predicted (i.e. $\hat{l}_n = 1$) over all negative trials. Varying $\theta$ gives a series of $\{\rm {TPR}(\theta), \rm {FPR}(\theta)\}$, which form an ROC curve as illustrated in Fig.\ref{fig:pAUC_schematic}. The gray area in Fig.\ref{fig:pAUC_schematic} illustrates how pAUC is defined. Specifically, it is defined as the area under the ROC curve when the value of FPR is between $[\alpha, \beta]$, where $\alpha$ and $\beta$ are two hyperparameters.
To calculate pAUC, we first construct two sets $\mathcal{P}=\{(s_i,l_i=1)|i=1,2,\cdots,I\}$ and $\mathcal{N}=\{(s_j,l_j=0)|j=1,2,\cdots,J\}$, where $I+J=N$. We then obtain a new subset $\mathcal{N}_0$ from $\mathcal{N}$ by adding the constraint $\rm{FPR}\in[\alpha, \beta]$ to $\mathcal{N}$ via following steps:
\begin{itemize}
\setlength{\itemsep}{3pt}
\setlength{\parsep}{1.5pt}
\setlength{\parskip}{1.5pt}
  \item[1)] $[\alpha$,$\beta]$ is replaced by $\left[j_\alpha/J,j_\beta/J \right]$ where $j_\alpha=\left \lceil J \alpha \right \rceil +1$ and $j_\beta=\left \lfloor J \beta \right \rfloor $ are two integers.
  \item[2)] $\{s_j\}_{\forall j: s_j\in \mathcal{N}}$ are sorted in descending order, where the operator $\forall a:b$ denotes that every $a$ that satisfies the condition $b$ will be included in the computation.
  \item[3)] $\mathcal{N}_0$ is selected as the set of the samples ranked from the top $j_\alpha \rm th$ to $j_\beta \rm th$ positions of the resorted $\{s_j\}_{\forall j: s_j\in \mathcal{N}}$, denoted as $\mathcal{N}_0=\{(s_k,l_k=0)|k=1,2,\cdots,K\}$ with $K=j_\beta - j_\alpha +1$.
\end{itemize}
Finally, pAUC is calculated as a normalized AUC over $\mathcal{P}$ and $\mathcal{N}_{0}$:
\begin{equation}
\label{eq:EmpAUC}
{\rm pAUC} = 1-\frac{1}{IK} \sum_{\forall i:s_i\in \mathcal{P}} \sum_{\forall k: s_k\in\mathcal{N}_0}\left[ \mathbb{I}(s_i<s_k ) + \frac{1}{2}\mathbb{I}(s_i= s_k)  \right]
\end{equation}
where  $\mathbb{I}(\cdot)$ is an indicator function that returns 1 if the statement is true, and 0 otherwise.
However, directly optimizing \eqref{eq:EmpAUC} is NP-hard. A common way to overcome the NP-hard problem is to relax the indicator function by a hinge loss function \cite{bai2019partial}:
\begin{equation}
\label{eq:hing}
    \ell_{\rm{hinge}}(z)={\rm{max}}(0, \delta-z)
\end{equation}
where $z=s_i-s_k$, and $\delta >0 $ is a tunable hyper-parameter. Because the gradient of \eqref{eq:hing} is a constant with respect to $z$, it does not reflect the difference between two samples that cause different errors. Motivated by the loss function of the least-squares support vector machine,
we replace \eqref{eq:hing} by \eqref{eq:hing1},
\begin{equation}
\label{eq:hing1}
    \ell_{\rm{hinge}}^\prime(z)={\rm{max}}(0, \delta-z)^2
\end{equation}
Substituting \eqref{eq:hing1} into \eqref{eq:EmpAUC} and changing the maximization problem \eqref{eq:EmpAUC} into an equivalent minimization, one can derive the following pAUC optimization objective:
\begin{equation}
\label{eq:loss}
     \min\frac{1}{IK}\sum_{\forall i:s_i\in \mathcal{P}}\sum_{\forall k: s_k\in\mathcal{N}_0} {\rm max}\left(0, \delta-(s_i-s_k)\right)^2
\end{equation}
The minimization of \eqref{eq:loss} needs to define a similarity function $f$ as illustrated in \eqref{eq:similarity}. We adopt the cosine similarity:
\begin{equation}\label{eq:cosine}
  s_n=f(\mathbf{x}_n,\mathbf{y}_n)=\frac{\mathbf{x}_n^T\mathbf{y}_n}{||\mathbf{x}_n||||\mathbf{y}_n||}
\end{equation}
where $||\cdot ||$ is the $\ell_2$-norm operator.

\subsection{Pairwise training set construction}\label{sec:pair}

Suppose that we have a training set $\mathcal{X}=\{\mathbf{x}_{uv}|u=1,\cdots,U;v=1,\cdots,V_u\}$, where $u$ and $v$ represent the $v_{th}$ utterance of the $u_{th}$ speaker, $U$ is the total number of the speakers and $V_u$ is the utterance number of the $u_{th}$  speaker.
If we construct $\mathcal{T}$ by using all of the above $\sum_{u=1}^{U}V_u$ utterances, the size of $\mathcal{T}$ would be enormous. So, in this work we propose the following two kinds of methods to construct the training set.

\textbf{\textit{1) Random sampling:}} We construct a set $\mathcal{T}^t$ at each mini-batch iteration of the DNN training by a random sampling strategy as follows. We first randomly select $t$ speakers from $\mathcal{X}$, then randomly select two utterances from each of the selected speakers, and finally construct $\mathcal{T}^t$ by a full permutation of the $2t$ utterances. It is easy to see that $\mathcal{T}^t$ contains $t$ true training trials and $[t(2t-1)-t]$ imposter training trials.

\textbf{\textit{2) Class-center learning:}} We construct a set $\mathcal{T}^{t_1}$ at each mini-batch iteration of the DNN training by a class-center learning algorithm as follows. Motivated from the identification loss, which learns a class center for each speaker, we assign a class center $\mathbf{w}$ to each speaker, denoted as $\{\mathbf{w}_u\}_{u=1}^U$.
At each iteration, we first select $t_1$ utterances randomly, and then combine them with $\{\mathbf{w}_u\}_{u=1}^U$ in pairwise to form $\mathcal{T}^{t_1}$, which contains $t_1$ true training trials and $(t_1U-t_1)$ imposter training trials. In the training stage, $\{\mathbf{w}_u\}_{u=1}^U$ is  randomly initialized and  updated at each iteration by the back propagation.

In comparison with the random sampling strategy, the class-center learning algorithm aggregates all training utterances of a speaker to its class-center. The class-centers should be more discriminative and robust than the random samples. Hence, training on $\mathcal{T}^{t_1}$ should be easier and more consistent than training on $\mathcal{T}^{t}$.
Note that the class-center learning algorithm requires the ground-truth speaker identities in training, which is same as the identification loss based speaker verification methods, because the class-centers rely on the ground-truth speaker identities. So, this algorithm is not applicable if the training labels are speaker trials instead of speaker identities.

\section{Connections to other loss functions }\label{sec:connections}
\subsection{Connection to cross-entropy minimization with softmax}
The softmax classifier is presented as
\begin{equation}\label{eq:softmax_classifier}
  \mathcal{L}_{\rm softmax}=-\frac{1}{R}\sum_{r=1}^{R}{\rm log}\frac{e^{\mathbf{w}_{y_r}^{\rm T}\mathbf{x}_r+b_{y_r}}}{\sum_{u=1}^{U}e^{\mathbf{w}_u^{\rm T}\mathbf{x}_r+b_u}},
\end{equation}
where $\mathbf{x}_r$ represents the $r$-th training sample, $y_r$ is its ground truth speaker identity, $\mathbf{w}_u$ is the $u$-th column of the weights of the output layer, $b_u$ is the bias term, and $R$ is the total number of the training samples.

The proposed method has a close connection to the softmax classifier on the class-center learning method. The objective function \eqref{eq:loss} aims to maximize the pAUC of the pairwise training set $\mathcal{T}^{t_1}$ at a mini-batch iteration, while the cross-entropy minimization with softmax aims to classify the $t_1$ utterances that are used to construct the $\mathcal{T}^{t_1}$. The class centers $\{\mathbf{w}_u\}_{u=1}^U$ are used for constructing $\mathcal{T}^{t_1}$ in the pAUC optimization, and used as the parameters of the softmax classifier in \eqref{eq:softmax_classifier}.

\subsection{Connection to triplet loss}
Triplet loss requires that the utterances from the same speaker are closer than those from different speakers in a triplet trial  \cite{li2017deep1}, i.e.,
\begin{equation}\label{eq:trip}
  f(\mathbf{x}^a,\mathbf{x}^p) - f(\mathbf{x}^a,\mathbf{x}^n) >  \sigma
\end{equation}
where $\mathbf{x}^a$, $\mathbf{x}^p$, and $\mathbf{x}^n$ represent the anchor, positive, and negative samples respectively, and $\sigma$ is a tunable hyperparameter.

The difference between pAUC and triplet loss lies in the following two aspects. First, according to \eqref{eq:EmpAUC} and \eqref{eq:hing}, the  relative constraint of pAUC can be written as
\begin{equation}
 s_i - s_k>  \delta
\end{equation}
where  $s_i$ and $s_k$ are constructed by four utterances. In other words, the relative constraint of pAUC is tetrad, which matches the pipeline of  speaker verification, while the relative constraint \eqref{eq:trip} is triplet. Second, pAUC is able to pick difficult training trials from the exponentially large number of training trials during the training process, while the triplet loss lacks such an ability.

\subsection{Connection to AUC maximization}
The AUC optimization \cite{mingote2019optimization} is a special case of the pAUC optimization with $\alpha=0$ and $\beta=1$. It is known that the performance of a speaker verification system is determined on the discriminability of the difficult trials. However, the AUC optimization is trained on $\mathcal{P}$ and $\mathcal{N}$. The two sets may contain many easy trials, which hinders the focus of the AUC optimization on solving the difficult problem. In comparison, the pAUC optimization with a small $\beta$ is able to select difficult trials at each mini-batch iteration. Experimental results in the following Section also demonstrate that the pAUC optimization is more effective than the AUC optimization.

\section{Experiments}\label{sec:exp}

\subsection{Data sets}
\label{sec:typestyle}
We conducted two experiments with the kaldi recipes \cite{povey2011kaldi} of \textit{``/egs/sitw/v2''}  and \textit{``/egs/sre16/v2''} respectively. Because the sampling rates of the training data of the first and second recipes are 16 kHz and 8 kHz respectively, we name the two experiments as the \textit{16KHZ system} and \textit{8KHZ system} accordingly for simplicity.
In the 16KHZ system, the deep embedding models were trained using the speech data extracted from the combined VoxCeleb 1 and 2 corpora \cite{nagrani2017voxceleb,Chung2018}. The back-ends were trained on a subset of the augmented VoxCeleb data, which contains 200000 utterances. The evaluation was conducted on the Speakers in the Wild (SITW) \cite{mclaren2016speakers} dataset, which has two evaluation tasks--Dev.Core and Eval.Core. In the 8KHZ system, the training data for the deep embedding models consist of Switchboard Cellular 1 and 2, Switchboard 2 Phase 1, 2, and 3,  NIST SREs from 2004 to 2010, and Mixer 6. The back-ends were trained with the NIST SREs along with Mixer 6. The evaluation data is the Cantonese language of NIST SRE 2016. We followed kaldi to train the PLDA adaptation model via the unlabeled data of NIST SRE 2016.

\subsection{Experimental setup}

We compared five loss functions, which are the the cross-entropy loss with softmax (Softmax) and additive angular margin softmax (ArcSoftmax) \cite{liu2019speaker}, random sampling based pAUC optimization (pAUC-R), class-center learning based pAUC optimization (pAUC-L), and  class-center learning  based AUC optimization (AUC-L), respectively. Besides, we also cited the published results in the kaldi source code, denoted as Softmax (kaldi), for comparison.

 We followed kaldi for the data preparation including the MFCC extraction, voice activity detection, and cepstral mean normalization.
For all comparison methods, the deep embedding models were trained with the same data augmentation strategy and DNN structure (except the output layer) as those in  \cite{snyder2018xvector}. They were implemented by Pytorch with the Adam optimizer. The learning rate was set to 0.001 without learning rate decay and weight decay. The batch-size was set to 128, except for pAUC-R whose batch-size was set to 512. The deep embedding models in the 16KHZ and 8 KHZ systems were trained with 50 and 300 epochs respectively. We adopted the LDA+PLDA back-end for all comparison methods. The dimension of LDA was set to 256 for the pAUC-L, AUC-L and ArcSoftmax of the 16KHZ system, and was set to 128 for the other evaluations.

For pAUC-R, the hyperparameter $\alpha$ was fixed to 0; the hyperparameter $\beta$ was set to 0.01 for the 16KHZ system and $0.1$ for the 8KHZ system; the hyperparameter $\delta$ was set to 1.2 for the 16KHZ system and 0.4 for the 8KHZ system;. For pAUC-L, $\alpha$ and $\delta$ were set the same as those of pAUC-R; $\beta$ was set to 0.001 for the 16KHZ system and 0.01 for the 8KHZ system. For ArcSoftmax, we adopted the best hyperparameter setting as that in \cite{liu2019speaker}.

The evaluation metrics include the equal error rate (EER), minimum detection cost function with $P_{\mathrm{target}} = 10^{-2}$ (DCF$10^{-2}$) and  $P_{\mathrm{target}} = 10^{-3}$ (DCF$10^{-3}$) respectively, and detection error tradeoff (DET) curve.
\subsection{Main results}
\label{sec:majhead}

\begin{table}[t!]
  \centering
  \caption{Results on SITW.}
  \label{tab:sitw}
   \scalebox{0.85}{
    \begin{threeparttable}
  \begin{tabular}{l|l|c|c|c}
    \hline
    \hline
    Name& Loss &$\rm{EER (\%)}$ &$\rm{DCF10^{-2}}$& $\rm{DCF10^{-3}}$\\
    \hline
   \multirow{6}{*}{Dev.Core}
   &Softmax (kaldi)            & 3.0          & -              &   -     \\
   \cline{2-5}
   &Softmax          &3.04          &0.2764          &\textbf{0.4349}  \\
   \cline{2-5}
   &ArcSoftmax       &\textbf{2.16} &\textbf{0.2565} & 0.4501  \\
   \cline{2-5}
   &pAUC-R           & 3.20         & 0.3412         & 0.5399          \\
   \cline{2-5}
   &pAUC-L           &\textbf{2.23} &\textbf{0.2523} & \textbf{0.4320}   \\
   \cline{2-5}
   &AUC-L            & 4.27        & 0.4474          & 0.6653       \\
   \hline
   \multirow{6}{*}{Eval.Core}
   &Softmax (kaldi)            & 3.5          & -              &-       \\
   \cline{2-5}
   &Softmax     &3.45          &0.3339          &\textbf{0.4898}  \\
   \cline{2-5}
   &ArcSoftmax      &\textbf{2.54} &\textbf{0.3025} &0.5142  \\
   \cline{2-5}
   &pAUC-R          & 3.74         &0.3880          & 0.5797       \\
   \cline{2-5}
   \cline{2-5}
   &pAUC-L          &\textbf{2.56} &\textbf{0.2949}& 0.5011  \\
   \cline{2-5}
   &AUC-L           &4.76          & 0.5005        & 0.7155        \\
   \hline
   \hline
  \end{tabular}
  \end{threeparttable}
}
\end{table}

The experimental results on SITW and NIST SRE 2016 are listed in Tables \ref{tab:sitw} and \ref{tab:sre16} respectively. From the results of Softmax, one can see that our implementation of Softmax via Pytorch achieves similar performance with the kaldi's implementation, which validates the correctness of our deep embedding model. We also observed that, if the stochastic gradient descent algorithm was carefully tuned with suitable weight decay, the performance can be further improved, which will not be reported in this paper due to the length limitation. Moreover, ArcSoftmax significantly outperforms Softmax, which corroborates the results in \cite{xie2019utterance,liu2019speaker}.

\begin{figure}[t!]
  \centering
  \includegraphics[width=2.8 in]{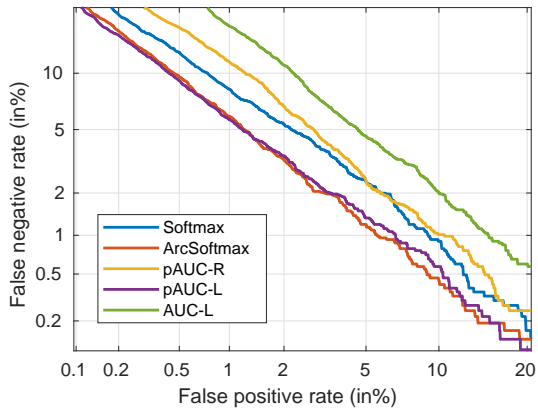}
  \caption{DET curves of the comparison methods on the Eval.Core. task.}
  \label{fig:det_curve}
\end{figure}

\begin{table}[t!]
  \centering
  \caption{Results on the Cantonese language of NIST SRE 2016.}
  \label{tab:sre16}
   \scalebox{0.82}{
  \begin{tabular}{l|l|c|c|c}
    \hline
    \hline
    Back-end & Loss &$\rm{EER (\%)}$ &$\rm{DCF10^{-2}}$& $\rm{DCF10^{-3}}$\\
    \hline
   \multirow{6}{*}{No-adaptation}
   &Softmax (kaldi)         &  7.52        &  -            &  -      \\
   \cline{2-5}
   &Softmax      &  6.76        & 0.5195        &0.7096    \\
   \cline{2-5}
   &ArcSoftmax   & \textbf{5.59}&\textbf{0.4640}&\textbf{0.6660}    \\
   \cline{2-5}
   &pAUC-R       & 15.25        &0.8397          & 0.9542          \\
   \cline{2-5}
   &pAUC-L       &\textbf{6.01} &\textbf{0.5026}&\textbf{0.7020}     \\
   \cline{2-5}
   &AUC-L        & 7.92         &0.5990         &0.8072                \\
   \hline
   \multirow{6}{*}{Adaptation}
   &Softmax (kaldi)        & 4.89        &  -            &  -   \\
   \cline{2-5}
   &Softmax     &4.94         & 0.4029        & 0.5949      \\
   \cline{2-5}
   &ArcSoftmax  &\textbf{4.13}&\textbf{0.3564}&\textbf{0.5401}    \\
   \cline{2-5}
   &pAUC-R      & 8.65        &0.6653         &0.8715        \\
   \cline{2-5}
   &pAUC-L      &\textbf{4.25}&\textbf{0.3704}&\textbf{0.5471}     \\
    \cline{2-5}
   &AUC-L       & 5.36        &0.4439         &0.6480      \\
   \hline
   \hline
  \end{tabular}
  }
\end{table}

 pAUC-L reaches an EER score of over $25\%$ and $10\%$ relatively lower than Softmax in the two experimental systems respectively. It also achieves comparable performance to the state-of-the-art ArcSoftmax, which demonstrates that the verification loss functions are comparable to the identification loss functions in performance. pAUC-L also outperforms pAUC-R significantly, which demonstrates that the class-center learning algorithm is a better training set construction method than the random sampling strategy. It is also seen that AUC-L cannot reach the state-of-the-art performance.
 The DET curves of the comparison methods are plotted in Fig. \ref{fig:det_curve}. From the figure, we observe that the DET curve of pAUC-L is close to that of ArcSoftmax, both of which perform the best among the studied methods.

\subsection{Effects of hyperparameters on performance}
\label{ssec:subhead}
This subsection investigates the effects of the hyperparameters of pAUC-L on performance. The hyperparameters were selected via $\alpha=0$, $\beta=(0,1]$, and $\delta=[0, 2)$. To accelerate the evaluation, we trained a pAUC-L model with 50 epochs using one quarter of the training data at each hyperparameter setting in the 16KHZ system. The evaluation results are listed in Table \ref{tab:par_sitw}. From the table, one can see that the parameter $\beta$, which controls the range of FPR for the pAUC-L optimization, plays a significant role on the performance. The performance is stable if $\beta\leq 0.1$, and drops significantly when $\beta=1$, i.e., the AUC-L case. This is because that pAUC-L focuses on discriminating the difficult trials automatically instead of considering all training trials as AUC-L did. It is also observed that the performance with the margin $\delta \geq 0.4$ is much better than that with $\delta=0$. We also evaluated pAUC-L in the 8KHZ system where the models were trained with 100 epochs using half of the training data. The results are presented in Table~\ref{tab:par_sre16}, which exhibits  the similar phenomena as in Table \ref{tab:par_sitw}.

Comparing Tables \ref{tab:par_sitw} and \ref{tab:par_sre16}, one may see that the optimal values of $\beta$ in the two evaluation systems are different. This is mainly due to the different difficulty levels of the two evaluation tasks. Specifically, the classification accuracies on the training data of the 16KHZ and 8KHZ systems are 97\% and 85\% respectively, which indicates that the training trials of the 16KHZ system are much easier to classify than the training trials of the 8KHZ system. Because the main job of $\beta$ is to select the training trials that are most difficult to be discriminated, setting $\beta$ in the 16KHZ system to a smaller value than that in the 8KHZ system helps both of the systems reach a balance between the problem of selecting the most difficult trials and gathering enough number of training trials for the DNN training.

\begin{table}[t!]
   \caption{EER (\%) of pAUC-L on the SITW DEV.Core task, where the keyword ``NaN'' means that the optimization process encounters numerical problems. }
   \centering
   \label{tab:par_sitw}
   \scalebox{0.88} {
\begin{tabular}{l|c|c|c|c|c }
  \hline
  \hline
    & $\delta=0.0$   & $\delta=0.4$ & $\delta=0.8$ & $\delta=1.2$ & $\delta=1.6$\\
  \hline
   $\beta=0.0001$  & -     & NaN  & -     &  -   & -   \\
  \hline
   $\beta=0.001$   & 4.69  & 3.04 & 2.71  & 2.58 & 2.81 \\
  \hline
   $\beta=0.01$    & 4.57  & 3.17 & 2.93  & 3.00 & 2.81 \\
  \hline
  $\beta=0.1$      &  -    & 3.14 &  -    & -    &  -  \\
  \hline
  $\beta=1$        &  -    & 4.12 &  -    &  -   &  -   \\
  \hline
  \hline
\end{tabular}
}
\end{table}

\begin{table}[t!]
   \caption{EER (\%) of pAUC-L on the Cantonese language. }
   \centering
   \label{tab:par_sre16}
   \scalebox{0.88} {
\begin{tabular}{l|c|c|c|c|c }
  \hline
  \hline
    & $\delta=0.0$   & $\delta=0.4$ & $\delta=0.8$ & $\delta=1.2$ & $\delta=1.6$\\
  \hline
   $\beta=0.001$   &  24.07 & 8.29 &  9.70 & 9.58 &   10.85   \\
  \hline
   $\beta=0.01$    &  11.74 & 7.40 &  7.52 & 7.64 &   7.38   \\
  \hline
  $\beta=0.1$      &  12.57 & 8.54 &  9.07 & 9.30 &   9.94   \\
  \hline
  \hline
\end{tabular}
}
\end{table}

\section{Conclusions}
\label{sec:conclusions}
This paper presented a method to train deep embedding based text-independent speaker verification with a new verification loss function---pAUC.
The major contribution of this paper consists of the following three respects. 1) A pAUC based loss function is proposed for deep embedding.
2) A method is presented to learn the class-centers of the training speakers for the training set construction. 3) we analyzed the connection between pAUC and the representative loss functions. The experimental results demonstrated that the proposed loss function is comparable to the state-of-the-art identification loss functions in speaker verification performance.

\small
\bibliographystyle{IEEEtran}
\bibliography{refs}

\end{document}